\documentclass[acmsmall,screen,nonacm]{acmart}
\usepackage{algorithm}
\usepackage{algpseudocode}
\usepackage{multirow}
\usepackage{adjustbox}
\newcommand{\Design}{\textsc{VikPath}}
\AtBeginDocument{%
  }

\begin{document}

\title{VikPath: A Vision Kansformer Framework for Effective Obstacle Avoidance in Self-Supervised Pathfinding}

\author{Junyao Wang}
\affiliation{%
  \department{Department of Computer Science}
  \institution{University of California, Irvine}
  \city{Irvine}
  \state{CA}
  \country{USA}}
\email{junyaow4@uci.edu}

\author{Yulin Xu}
\affiliation{%
  \department{Department of Electrical Engineering and Computer Science}
  \institution{University of California, Irvine}
  \city{Irvine}
  \state{CA}
  \country{USA}}
\email{yulinx8@uci.edu}

\author{Mohammad Abdullah Al Faruque}
\affiliation{%
  \department{Department of Electrical Engineering and Computer Science}
  \institution{University of California, Irvine}
  \city{Irvine}
  \state{CA}
  \country{USA}}
\email{alfaruqu@uci.edu}

\begin{abstract}
Pathfinding is a fundamental problem in artificial intelligence and autonomous systems. Traditional heuristic-based algorithms, such as A*, rely on predefined heuristic functions to guide the search process. Although effective in structured environments, their search efficiency can degrade substantially in complex, obstacle-rich scenarios, where handcrafted heuristics may provide limited guidance. Recent studies have explored learning-based approaches to improve pathfinding efficiency; however, most existing methods rely on supervised learning and require labels generated by conventional planners or obtained through manual annotation. As a result, their performance is inherently influenced by the quality of the underlying supervision and may degrade when the labeling heuristics fail to capture complex environmental structures. Moreover, existing methods primarily optimize for path length while paying limited attention to obstacle clearance and trajectory smoothness, which can lead to paths that are difficult or unsafe to execute in real-world environments. 
To address these limitations, we propose $\Design$, a self-supervised pathfinding framework that jointly considers obstacle proximity and path smoothness. At its core, our novel \textit{Vision Kansformer} module learns representations of obstacle distributions without relying on labeled trajectories, enabling the model to better adapt to complex environments. We further introduce a sharp-turn penalty to encourage smoother and more practically executable paths. Extensive experiments demonstrate that, compared with state-of-the-art (SOTA) approaches, $\Design$ achieves an average of 3.28\% greater obstacle clearance and 87.07\% lower inference latency while maintaining smooth path generation.


\end{abstract}



\maketitle
\thispagestyle{empty}
\pagestyle{standardpagestyle}
\fancyfoot[RO,LE]{}
\section{Introduction}\label{sec:intro}

Pathfinding is a fundamental problem in artificial intelligence, robotics, and autonomous systems. It aims to determine an efficient and collision-free route for a mobile agent from a starting point to a destination~\cite{karur2021survey,qin2023review,wang2024rs2g}. Traditional heuristic-based algorithms, such as A*~\cite{hart1968formal,pohl1970heuristic}, typically represent the environment as a grid or graph and use heuristic-guided search to identify a shortest feasible path~\cite{foead2021systematic,wang2026cruise}. Despite their effectiveness, such representations often provide only limited descriptions of the environment and may fail to capture richer spatial structures that could facilitate efficient planning. Moreover, conventional heuristics are typically handcrafted and rely on relatively simple estimates of the remaining path cost, limiting their ability to exploit complex obstacle configurations and global environmental patterns~\cite{andreychuk2022multi}. This reliance on manually designed heuristics can also reduce adaptability across environments with diverse map layouts and obstacle distributions~\cite{pohl1970heuristic,ruml2007best}. More importantly, classical planners primarily optimize path length and collision avoidance, while often overlooking practical factors such as obstacle clearance and trajectory smoothness. As a result, the generated paths may pass unnecessarily close to obstacles or contain abrupt turns, making them less desirable for real-world navigation~\cite{dijkstra2022note,dolgov2010path}. These limitations have motivated increasing interest in learning-based approaches that leverage environmental structure to improve pathfinding performance~\cite{yonetani2021path,veerapaneni2023learning}.

\begin{figure}[!t]
\centering
\includegraphics[width=0.6\linewidth]{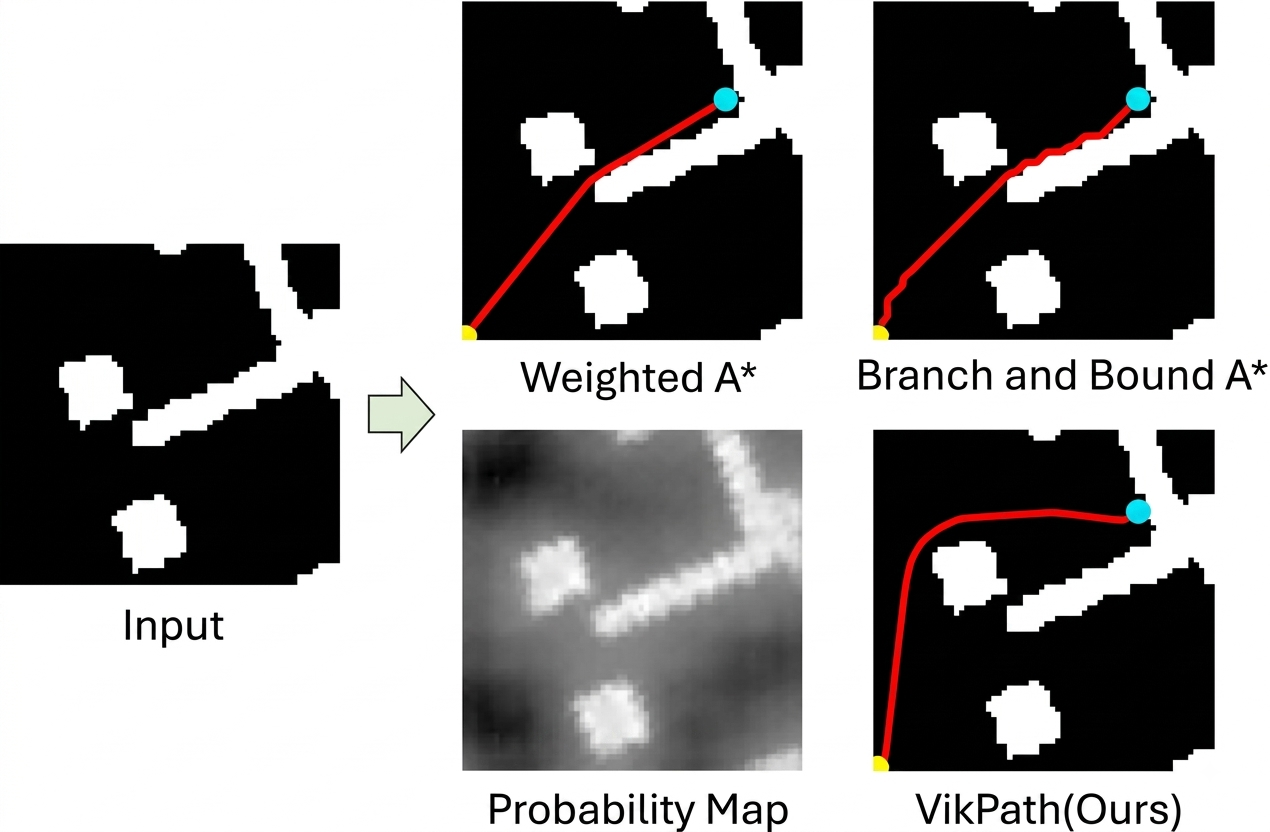}
\caption{Existing pathfinding algorithms, e.g., A* algorithm, do not consider obstacle proximity, resulting in paths that are impractically close to obstacles. In contrast, VikPath constructs probability maps to represent obstacle distributions, ensuring paths avoid areas with high obstacle density.}
\vspace{-2mm}
\label{fig:intro}  
\end{figure}

Learning-based approaches have been increasingly explored to improve pathfinding performance, with many supervised methods relying on expert paths generated by conventional planners or manually annotated trajectories as supervision~\cite{kirilenko2023transpath,yonetani2021path}. However, the effectiveness of these methods is inherently dependent on the quality of the supervisory signals, making learned planners susceptible to errors and biases present in the training labels~\cite{jiang2020supervised,cunningham2008supervised}. Moreover, planner-generated labels typically reflect the objectives and inductive biases of the underlying algorithms, which may limit the adaptability and generalization of learned models across environments with diverse obstacle configurations and spatial structures~\cite{almuqati2024challenges}. In particular, when supervision is primarily derived from shortest-path planners, learned models tend to prioritize path length while paying limited attention to practical criteria such as obstacle clearance and trajectory smoothness~\cite{yonetani2021path,kirilenko2023transpath,kim2026flexpath}. Recent approaches have sought to alleviate this dependence on expert-generated paths through self-supervised optimization or reinforcement learning (RL). For example, iA$^*$ introduces self-supervised guidance through differentiable search, while RL-based methods learn navigation policies through direct interactions with the environment~\cite{chen2025ia,hu2021path,aberdeen2005policy}. Although these approaches reduce the need for explicit path annotations, self-supervised search methods have primarily focused on improving search efficiency and path length, whereas RL-based methods often require extensive environment interactions and substantial computational resources for training~\cite{chen2025ia,liu2021policy}. More recently, FlexPath explores adapting learned planning guidance to alternative path preferences, including obstacle clearance, yet still relies on planner-generated demonstrations to learn an initial connectivity prior~\cite{kim2026flexpath}. In parallel, trajectory smoothness is commonly addressed through post-processing techniques, such as Bézier-curve-based smoothing, after an initial route has been generated~\cite{qiao2023end,cimurs2017bezier}. Because path generation and smoothing are optimized separately, such post-processing may alter the original trajectory, reduce obstacle clearance, or require additional feasibility checks in obstacle-rich environments. These limitations motivate a unified pathfinding framework that eliminates reliance on expert-generated path labels while explicitly incorporating obstacle clearance and trajectory smoothness into the path generation process.

To address these limitations, we propose $\Design$, an efficient self-supervised pathfinding framework that jointly considers obstacle proximity, path smoothness, and computational efficiency. Unlike existing supervised approaches that rely on heuristic-generated or manually annotated paths, $\Design$ learns spatial structures directly from unlabeled environment maps through a novel \textit{Vision Kansformer} module. By performing masked reconstruction, Vision Kansformer captures both local and global spatial dependencies and produces a fine-grained probability map that characterizes the obstacle distribution. This representation provides explicit obstacle-density awareness during planning, discouraging traversal through obstacle-dense regions and improving path clearance. We further augment the weighted A* cost function with obstacle-proximity awareness and a sharp-turn penalty, allowing path safety and smoothness to be considered directly during search and eliminating the need for a separate Bézier-curve-based smoothing stage. The main contributions of this paper are summarized as follows:
\begin{itemize}
\item We propose $\Design$, a self-supervised pathfinding framework that learns environmental representations without heuristic-generated or manually annotated path labels and jointly incorporates obstacle awareness and path smoothness into planning. $\Design$ achieves, on average, 87.07\% lower inference latency than state-of-the-art (SOTA) methods.

\item We introduce a novel \textit{Vision Kansformer} module that leverages masked modeling to capture both local and global spatial structures and construct a fine-grained representation of obstacle distributions. This obstacle-aware representation enables $\Design$ to generate paths with, on average, 3.28\% greater obstacle clearance than SOTA methods.

\item We enhance the weighted A* algorithm by explicitly incorporating obstacle proximity and a sharp-turn penalty into its cost function, enabling obstacle-aware and smooth path generation without requiring a separate Bézier-curve-based post-processing step.

\end{itemize}

\section{Related Works}\label{sec:related}

\subsection{Learning-based Pathfinding}

Traditional graph-search algorithms, such as Dijkstra's algorithm~\cite{dijkstra2022note} and A*~\cite{hart1968formal}, have long served as fundamental tools for pathfinding. While Dijkstra's algorithm searches according to accumulated path cost, A* improves search efficiency by incorporating a heuristic estimate of the remaining cost to the goal. However, commonly used heuristics, such as Euclidean or Manhattan distance, encode limited information about obstacle configurations and may provide insufficient guidance in complex, obstacle-rich environments~\cite{kirilenko2023transpath}. This limitation has motivated learning-based approaches that leverage environmental information to improve search guidance and planning efficiency~\cite{takahashi2019learning,numeroso2022learning}. Gated Path Planning Networks (GPPNs)~\cite{lee2018gated}, for example, formulate planning as recurrent convolutional computation and employ gated recurrent updates to propagate spatial information. Neural A*~\cite{yonetani2021path} combines a convolutional encoder with differentiable A* search to learn guidance maps from expert-generated paths, while TransPath~\cite{kirilenko2023transpath} employs convolutional and attention-based architectures to learn instance-dependent heuristic proxies for grid-based pathfinding. Despite their effectiveness, these supervised approaches rely on expert- or planner-generated targets, making their learned guidance dependent on the quality and objectives of the underlying supervision. Recent studies have therefore explored reducing such reliance on labeled paths. iA$^*$~\cite{chen2025ia} introduces a self-supervised path-planning framework that jointly optimizes learned search guidance and differentiable A* through bilevel optimization. Reinforcement learning (RL)-based approaches provide another alternative by learning navigation policies through direct interaction with the environment~\cite{panov2018grid,liu2024deep}; however, they typically require extensive environment interactions and substantial training resources. Beyond search efficiency and path length, recent work has also begun to consider alternative path-quality objectives. FlexPath~\cite{kim2026flexpath}, for instance, adapts a learned connectivity prior toward path preferences such as obstacle clearance, although its initial representation is learned from shortest-path demonstrations. DAA$^*$~\cite{xu2025daa} explicitly incorporates angular information into learned A* search to improve path smoothness, while still relying on expert-path supervision. Overall, jointly learning obstacle-aware environmental representations and directly incorporating both obstacle clearance and path smoothness into planning without expert-generated path labels remains relatively underexplored.

\subsection{Masked Modeling}

Masked image modeling (MIM) has emerged as a prominent paradigm for self-supervised visual representation learning, in which portions of an input are masked and a model is trained to recover the missing information from the visible context~\cite{li2023masked,hondru2025masked}. By requiring predictions from incomplete observations, MIM encourages models to capture contextual relationships and structural information without relying on manual annotations. Vision Transformer (ViT)~\cite{dosovitskiy2020image} provides an effective patch-based Transformer architecture upon which numerous masked modeling approaches have been developed. SiT~\cite{ahmed2021sit} introduces Group Masked Model Learning (GMML), which masks spatially connected groups of image patches and reconstructs them using their surrounding context. BEiT~\cite{bao2021beit} formulates masked image modeling as the prediction of discrete visual tokens, whereas Masked Autoencoder (MAE)~\cite{he2021masked} directly reconstructs randomly masked image patches using an asymmetric encoder-decoder architecture. SimMIM~\cite{xie2022simmim} further simplifies masked image modeling by adopting a lightweight prediction head and a direct pixel-level reconstruction objective. PeCo~\cite{dong2023peco} incorporates a perceptually informed visual tokenizer to encourage the reconstruction of semantically meaningful visual content. Collectively, these studies demonstrate the effectiveness of masked reconstruction for learning representations from unlabeled visual inputs. Inspired by the group masking strategy of SiT~\cite{ahmed2021sit}, our work applies structured masking to environmental maps and further tailors the reconstruction objective to capture both large-scale spatial structures and fine-grained obstacle distributions for pathfinding.

\subsection{Kolmogorov-Arnold Networks}

The Kolmogorov-Arnold representation theorem~\cite{kolmogorov1957representation} establishes that continuous multivariate functions can be represented through compositions and sums of continuous univariate functions. Inspired by this theorem, Kolmogorov-Arnold Networks (KANs)~\cite{liu2024kan} replace the fixed activation functions and scalar weights used in conventional multilayer perceptrons (MLPs) with learnable univariate functions, typically parameterized using B-splines. This formulation provides flexible function approximation and allows individual learned functions to be directly inspected and refined. The original KAN study reports favorable approximation scaling and interpretability on several scientific learning tasks~\cite{liu2024kan}, although subsequent studies indicate that these advantages do not universally translate to all domains when KANs and MLPs are compared under matched computational budgets~\cite{yu2024kan}. Consequently, recent research has focused on adapting KANs to different architectures and application domains. U-KAN~\cite{li2024u} integrates KAN layers into a U-Net-style architecture for medical image segmentation and generation, while Convolutional KANs extend KAN-style learnable functions to convolutional operators~\cite{dylan2024convolutional}. KAN-based architectures have also been investigated for time-series modeling and control~\cite{vaca2024kolmogorov}, reinforcement learning~\cite{kich2024kolmogorov}, and implicit neural representations~\cite{mehrabian2024implicit}. More closely related to our architecture, the Kolmogorov-Arnold Transformer (KAT)~\cite{yang2025kolmogorov} replaces conventional Transformer MLP blocks with KAN-based layers to improve nonlinear representation learning. Building on these developments, our \textit{Vision Kansformer} integrates KAN-based transformations into both the Transformer architecture and reconstruction head and couples them with masked modeling to learn obstacle-aware representations specifically for self-supervised pathfinding.

\begin{figure*}[t] 
\centering
    \includegraphics[width=\textwidth]{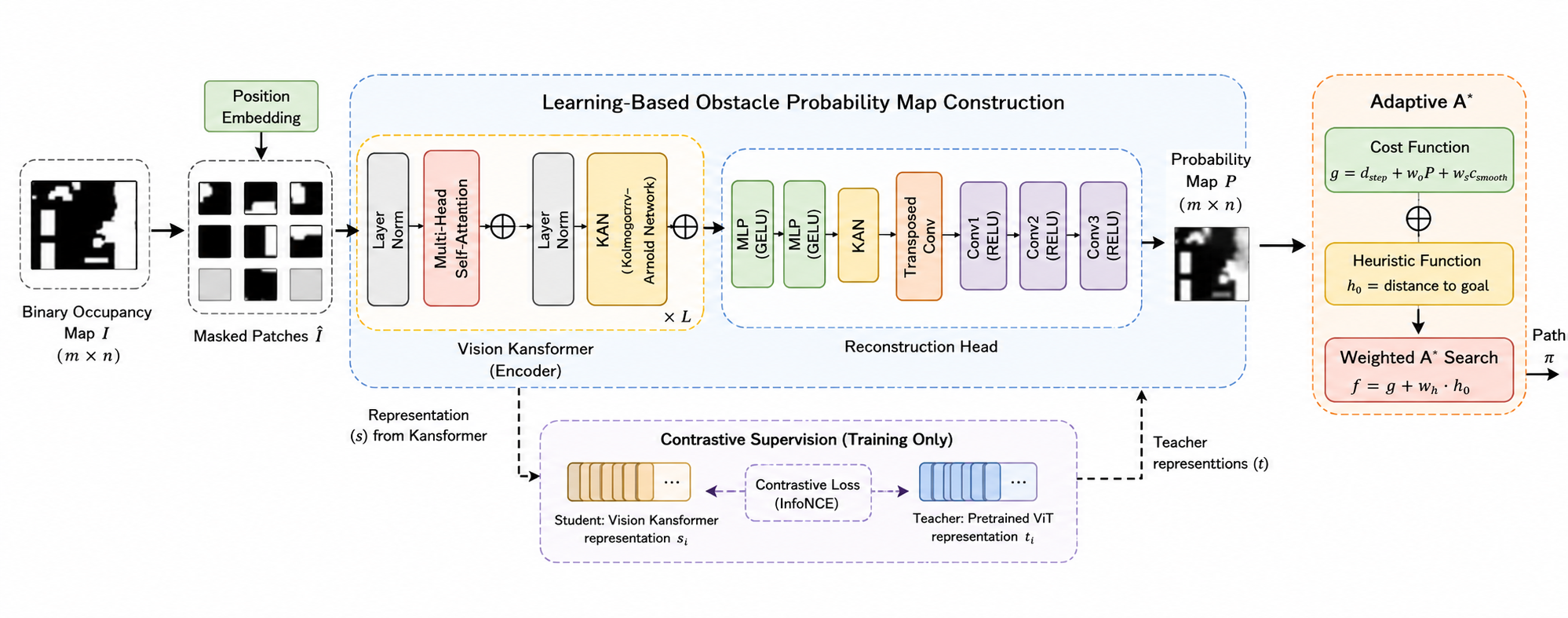}
   \caption{Overview of the $\Design$ architecture. The Vision Kansformer learns
obstacle-aware representations from masked occupancy maps under reconstruction,
perceptual, and contrastive supervision. A reconstruction head converts the
learned representations into an obstacle probability map, which is incorporated
into an obstacle- and smoothness-aware Weighted A* search to generate the final
path. The pre-trained ViT teacher is used only during training for contrastive
supervision.}
    \label{fig:workflow}
\end{figure*}
\section{Methodology}
\label{sec:method}

\subsection{Problem Formulation}
\label{subsec:problem_formulation}

An overview of $\Design$ is shown in Figure~\ref{fig:workflow}. Given an
environment map, a start location, and a goal location, our objective is to
identify a collision-free path that jointly considers path efficiency,
obstacle clearance, and trajectory smoothness. $\Design$ follows a two-stage
framework consisting of (1) learning an obstacle-aware probability map from
the environment and (2) performing obstacle- and smoothness-aware path search
using weighted A* (WA*). 
We represent the input environment as a binary occupancy map
$I\in\{0,1\}^{m\times n}$, where 
\begin{equation}
I_{i,j}
=
\begin{cases}
0, & \text{if location } (i,j) \text{ corresponds to free space},\\
1, & \text{if location } (i,j) \text{ corresponds to an obstacle}.
\end{cases}
\label{eq:occupancy_map}
\end{equation} 
Let $S$ and $G$ denote the start and goal locations, respectively. A feasible
path is represented as a sequence of adjacent grid locations 
\begin{equation}
\pi
=
(Q_0,Q_1,\ldots,Q_T),
\label{eq:path_sequence}
\end{equation} 
where $Q_0=S$, $Q_T=G$, and every location along the path lies in free space. 
Although the binary occupancy map explicitly indicates which locations are
occupied, it does not directly encode the broader spatial context surrounding
each free-space location, such as its proximity to nearby obstacles or the
distribution of obstacles in the surrounding region. To capture such
information, we employ a learning-based \textit{Vision Kansformer} module to
construct an obstacle probability map

\begin{equation}
P
\in
[0,1]^{m\times n}.
\label{eq:probability_map}
\end{equation}
Operationally, each $P_{i,j}$ is a normalized obstacle-related score learned
from the spatial context of the environment. Larger values indicate stronger
obstacle-related influence, whereas smaller values correspond to relatively
open regions. We use the term \emph{obstacle probability map} throughout the
paper to refer to this normalized spatial representation. 
Given $P$, $\Design$ performs path search from $S$ to $G$ using an
obstacle- and smoothness-aware WA* formulation. The cumulative path cost
jointly accounts for movement distance, obstacle proximity, and changes in
path direction, while the weighted heuristic provides goal-directed geometric
guidance. Together, these components encourage efficient paths with greater
obstacle clearance and smoother trajectories.

\subsection{Learning-based Obstacle Probability Map}
\label{subsec:probability_map}

The first stage of $\Design$ learns a dense obstacle-aware representation
directly from environment maps without requiring heuristic-generated or
manually annotated path labels. As illustrated in Figure~\ref{fig:workflow},
this stage consists of masked input construction, the \textit{Vision
Kansformer} encoder, complementary representation-learning objectives, and a
reconstruction head that generates the obstacle probability map.

\subsubsection{Masking}
\label{subsubsec:masking}

We adopt a masking-and-reconstruction strategy to encourage the model to infer
obstacle structures from spatial context. Given an occupancy map
$I\in\{0,1\}^{m\times n}$, we construct a binary mask
$M\in\{0,1\}^{m\times n}$ as 
\begin{equation}
M_{i,j}
=
\begin{cases}
0, & \text{if location } (i,j) \text{ is masked},\\
1, & \text{otherwise}.
\end{cases}
\label{eq:mask}
\end{equation} 
The masked input is obtained as
\begin{equation}
\hat{I}
=
I\odot M,
\label{eq:masked_input}
\end{equation} 
where $\odot$ denotes element-wise multiplication. We further define the set
of masked locations as

\begin{equation}
\Omega_M
=
\left\{
(i,j)\mid M_{i,j}=0
\right\}.
\label{eq:masked_set}
\end{equation} 
By reconstructing the masked regions from the remaining visible context, the
model is encouraged to capture spatial dependencies and contextual obstacle
structures rather than relying solely on individual occupancy values.

\subsubsection{Vision Kansformer}
\label{subsubsec:kansformer}

Our \textit{Vision Kansformer} combines the global context modeling capability
of self-attention with the flexible nonlinear transformations of
Kolmogorov-Arnold Networks (KANs)~\cite{liu2024kan}. As illustrated in
Figure~\ref{fig:workflow}, the masked occupancy map $\hat{I}$ is first divided
into $N$ non-overlapping patches and linearly projected into $D$-dimensional
patch embeddings

\[
\mathbf{X}
\in
\mathbb{R}^{N\times D}.
\] 
To preserve spatial information, positional embeddings
$\mathbf{E}_{\mathrm{pos}}\in\mathbb{R}^{N\times D}$ are added to the patch
representations. The initial representation fed into Vision Kansformer is

\begin{equation}
\mathbf{Z}_0
=
\mathbf{X}
+
\mathbf{E}_{\mathrm{pos}}.
\label{eq:initial_embedding}
\end{equation} 
Unlike a conventional Transformer block, which employs a feed-forward network
(FFN) after self-attention, Vision Kansformer replaces the FFN with a KAN
layer. For the $l$-th Vision Kansformer block, multi-head self-attention
(MHSA) is first applied to the normalized input: 
\begin{equation}
\mathbf{Z}'_l
=
\mathbf{Z}_{l-1}
+
\operatorname{MHSA}
\left(
\operatorname{LN}(\mathbf{Z}_{l-1})
\right),
\label{eq:mhsa}
\end{equation} 
followed by a KAN transformation with a residual connection: 
\begin{equation}
\mathbf{Z}_l
=
\mathbf{Z}'_l
+
\operatorname{KAN}
\left(
\operatorname{LN}(\mathbf{Z}'_l)
\right).
\label{eq:kan_block}
\end{equation} 
Here, $\operatorname{LN}(\cdot)$ denotes layer normalization. MHSA captures
long-range dependencies among spatially separated patches, while the KAN layer
provides flexible nonlinear transformations for refining the learned
representations. Together, these components enable Vision Kansformer to model
global environmental structures while retaining fine-grained information
about local obstacle configurations. 
Following the standard KAN formulation~\cite{liu2024kan}, each learnable
univariate function is represented as a combination of a base function and a
B-spline expansion: 
\begin{equation}
\phi(z)
=
w_{\mathrm{base}}\,b(z)
+
w_{\mathrm{spline}}
\sum_r c_r B_r(z),
\label{eq:kan_function}
\end{equation} 
where $b(\cdot)$ denotes the base function, $B_r(\cdot)$ is the $r$-th
B-spline basis function, $c_r$ is its learnable coefficient, and
$w_{\mathrm{base}}$ and $w_{\mathrm{spline}}$ control the contributions of
the base and spline components, respectively. The local support of B-spline
basis functions enables localized nonlinear transformations, while the spline
grid determines the resolution of the learned functions. 
To adapt the spline functions to different activation distributions, the grid
is updated by interpolating between a uniformly spaced reference grid
$\mathbf{g}_{\mathrm{uniform}}$ and a data-adaptive grid
$\mathbf{g}_{\mathrm{adaptive}}$: 
\begin{equation}
\mathbf{g}
=
\epsilon\,\mathbf{g}_{\mathrm{uniform}}
+
(1-\epsilon)\,\mathbf{g}_{\mathrm{adaptive}},
\label{eq:adaptive_grid}
\end{equation} 
where $\epsilon\in[0,1]$ controls the relative contribution of the uniform and
adaptive grids.

\subsubsection{Training Objectives}
\label{subsubsec:training_objectives}

We train the obstacle probability map construction module using three
complementary objectives: a contrastive loss
$\mathcal{L}_{\mathrm{contrastive}}$ for enhancing representation
discriminability, a reconstruction loss
$\mathcal{L}_{\mathrm{reconstruction}}$ for learning
obstacle-proximity-aware spatial structures, and a perceptual loss
$\mathcal{L}_{\mathrm{perceptual}}$ for preserving higher-level structural
information. The overall training objective is 
\begin{equation}
\mathcal{L}
=
\beta\,\mathcal{L}_{\mathrm{contrastive}}
+
\lambda\,\mathcal{L}_{\mathrm{reconstruction}}
+
\alpha\,\mathcal{L}_{\mathrm{perceptual}},
\label{eq:total_loss}
\end{equation} 
where $\beta$, $\lambda$, and $\alpha$ are weighting coefficients that balance
the contributions of the three objectives.

\textbf{Contrastive Loss.}
To provide representation-level supervision during training, we employ a
Vision Transformer (ViT)~\cite{dosovitskiy2020image} pre-trained on
ImageNet~\cite{deng2009imagenet} as a teacher network. The teacher provides
reference representations for the contrastive objective but is not used as an
additional input to Vision Kansformer. 
Let $\mathbf{s}_i$ denote the representation produced by Vision Kansformer for
sample $i$, and let $\mathbf{t}_i$ denote the corresponding representation
extracted by the teacher network. For a mini-batch containing $N_b$ samples,
the contrastive loss is defined as 
\begin{equation}
\mathcal{L}_{\mathrm{contrastive}}
=
-\frac{1}{N_b}
\sum_{i=1}^{N_b}
\log
\frac{
\exp\left(
\operatorname{sim}(\mathbf{s}_i,\mathbf{t}_i)/\tau
\right)
}{
\sum_{k=1}^{N_b}
\exp\left(
\operatorname{sim}(\mathbf{s}_i,\mathbf{t}_k)/\tau
\right)
}.
\label{eq:contrastive_loss}
\end{equation} 
Here, $\operatorname{sim}(\cdot,\cdot)$ denotes cosine similarity and $\tau$
is a temperature parameter. The pair $(\mathbf{s}_i,\mathbf{t}_i)$ forms a
positive pair corresponding to the same environment, whereas teacher
representations from other samples serve as negatives. This objective
encourages Vision Kansformer to learn discriminative environment
representations while maintaining consistency with the teacher representation
space.

\textbf{Reconstruction Loss.}
The original binary occupancy map explicitly indicates whether each location
is occupied, but it does not directly characterize the spatial influence of
nearby obstacles. To construct an obstacle-proximity-aware reconstruction
target, we iteratively expand the obstacle regions in $I$. 
Let $O^{(0)}=I$ denote the original binary obstacle map. We recursively expand
the obstacle region for $L$ iterations as 
\begin{equation}
O^{(\ell)}
=
\mathcal{D}_{\mathcal{K}}
\left(
O^{(\ell-1)}
\right),
\qquad
\ell=1,\ldots,L,
\label{eq:iterative_expansion}
\end{equation} 
where $\mathcal{D}_{\mathcal{K}}(\cdot)$ denotes binary morphological dilation
with structuring element $\mathcal{K}$. Successive dilation steps
progressively include free-space locations farther from the original obstacle
boundary. 
To distinguish regions introduced at different expansion steps, we define the
$\ell$-th expansion ring as 
\begin{equation}
A^{(\ell)}
=
\begin{cases}
O^{(0)}, & \ell=0,\\[2mm]
O^{(\ell)}-O^{(\ell-1)}, & \ell=1,\ldots,L.
\end{cases}
\label{eq:expansion_ring}
\end{equation} 
We assign a monotonically decreasing proximity weight $\rho_\ell$ to each
expansion ring: 
\begin{equation}
1
=
\rho_0
\geq
\rho_1
\geq
\cdots
\geq
\rho_L
\geq
0.
\label{eq:proximity_weights}
\end{equation} 
The resulting obstacle-proximity-aware reconstruction target is 
\begin{equation}
R(I)_{i,j}
=
\sum_{\ell=0}^{L}
\rho_\ell
A^{(\ell)}_{i,j}.
\label{eq:pseudo_label}
\end{equation} 
Accordingly, obstacle locations receive the largest target values, while
free-space locations introduced in later expansion steps receive progressively
smaller values. Locations outside the $L$-step expanded obstacle region are
assigned zero. In this way, $R(I)$ encodes both binary obstacle occupancy and
the spatial proximity of free-space locations to nearby obstacles. 
Let $\tilde{I}$ denote the reconstructed obstacle-aware output predicted by
the network. The reconstruction loss is computed only over the masked
locations: 
\begin{equation}
\mathcal{L}_{\mathrm{reconstruction}}
=
\frac{1}{|\Omega_M|}
\sum_{(i,j)\in\Omega_M}
\operatorname{BCE}
\left(
\tilde{I}_{i,j},
R(I)_{i,j}
\right).
\label{eq:reconstruction_loss}
\end{equation} 
Here, $\Omega_M$ is defined in Equation~\ref{eq:masked_set}. By evaluating
the reconstruction objective only over masked regions, the model must infer
obstacle-related spatial information from the surrounding visible context.
The proximity-aware target $R(I)$ further encourages the model to distinguish
open regions from regions close to obstacles.

\textbf{Perceptual Loss.}
We further employ a perceptual loss to preserve higher-level structural
information during reconstruction. Specifically, we use a VGG16
network~\cite{simonyan2014very} pre-trained on
ImageNet~\cite{deng2009imagenet}. Let $\phi_l(\cdot)$ denote the feature
representation extracted from the $l$-th selected VGG16 layer. The perceptual
loss is defined as 
\begin{equation}
\mathcal{L}_{\mathrm{perceptual}}
=
\sum_l
\left\|
\phi_l(I)
-
\phi_l(\tilde{I})
\right\|_2.
\label{eq:perceptual_loss}
\end{equation} 
The reconstruction and perceptual objectives play complementary roles. The
reconstruction loss drives the prediction toward the
obstacle-proximity-aware target $R(I)$, whereas the perceptual loss regularizes
the reconstruction to preserve the higher-level structural characteristics of
the original occupancy map $I$.

\subsubsection{Probability Map Generation}
\label{subsubsec:probability_map_generation} 
As illustrated in Figure~\ref{fig:workflow}, the reconstruction head
transforms the representations learned by Vision Kansformer into a dense
obstacle probability map. Let $\mathbf{H}$ denote the output representation
of the final Vision Kansformer block. The reconstruction head first processes
$\mathbf{H}$ using two multilayer perceptron (MLP) layers, each followed by a
GELU activation. Let $\mathbf{H}_{\mathrm{mlp}}$ denote the resulting
representation. 
The representation is further refined by a KAN layer with a residual
connection: 
\begin{equation}
\mathbf{H}_{\mathrm{r}}
=
\mathbf{H}_{\mathrm{mlp}}
+
\operatorname{KAN}
\left(
\mathbf{H}_{\mathrm{mlp}}
\right).
\label{eq:reconstruction_kan}
\end{equation} 
The refined representation $\mathbf{H}_{\mathrm{r}}$ is reshaped into a
spatial feature map and passed through a transposed convolution to recover
spatial resolution. The upsampled representation is subsequently processed by
three convolutional layers to progressively refine the spatial features. Let
$\mathbf{F}\in\mathbb{R}^{m\times n\times C}$ denote the resulting
$C$-channel feature map. 
We aggregate the feature channels into a single spatial score map as 
\begin{equation}
A_{i,j}
=
\frac{1}{C}
\sum_{c=1}^{C}
\mathbf{F}_{i,j,c}.
\label{eq:channel_average}
\end{equation} 
Finally, min--max normalization is applied to obtain the obstacle probability
map: 
\begin{equation}
P_{i,j}
=
\frac{
A_{i,j}-A_{\min}
}{
A_{\max}-A_{\min}+\delta
},
\label{eq:probability_normalization}
\end{equation} 
where $A_{\min}=\min_{u,v}A_{u,v}$,
$A_{\max}=\max_{u,v}A_{u,v}$, and $\delta>0$ is a small constant introduced
for numerical stability. 
The resulting map $P\in[0,1]^{m\times n}$ represents a normalized
obstacle-related score over the environment. Since the reconstruction target
assigns larger values to obstacles and their nearby regions, larger values of
$P_{i,j}$ indicate stronger obstacle-related influence at location $(i,j)$,
whereas smaller values correspond to relatively open regions. This map is
subsequently incorporated into the path search to guide the planner away from
regions with strong obstacle influence.

\subsection{Obstacle- and Smoothness-Aware Weighted A* Pathfinding}
\label{subsec:weighted_astar}

Given the learned obstacle probability map $P$, the second stage of $\Design$
performs search-based pathfinding from the start location $S$ to the goal
location $G$. Rather than optimizing traversal distance alone, our formulation
augments weighted A* with an obstacle-related traversal penalty derived from
$P$ and a turning penalty that discourages abrupt changes in direction. 
Because the smoothness cost depends on the direction from which a location is
reached, representing a search state solely by its current location is
insufficient. We therefore define each search state as 
\begin{equation}
X
=
(Q_p,Q),
\label{eq:search_state}
\end{equation} 
where $Q$ is the current location and $Q_p$ is its predecessor along the
current partial path. The initial state is
$X_S=(\varnothing,S)$. States that arrive at the same location through
different predecessors are treated as distinct, since they may incur
different turning costs in subsequent transitions.

\subsubsection{Weighted Heuristic}
\label{subsubsec:weighted_heuristic}

Let $h_0(Q,G)$ denote a geometric estimate of the remaining travel cost from
the current location $Q$ to the goal $G$. For a 4-connected grid, we use the
Manhattan distance 
\begin{equation}
h_0(Q,G)
=
|Q_x-G_x|
+
|Q_y-G_y|,
\label{eq:base_heuristic}
\end{equation} 
where $(Q_x,Q_y)$ and $(G_x,G_y)$ denote the coordinates of $Q$ and $G$,
respectively. If an 8-connected neighborhood is used, the corresponding
diagonal-aware distance can be adopted instead. 
Following weighted A*, the priority of a search state $X=(Q_p,Q)$ is defined
as 
\begin{equation}
f(X)
=
g(X)
+
w_h\,h_0(Q,G),
\label{eq:weighted_total_cost}
\end{equation} 
where $g(X)$ is the cumulative cost of the partial path from $S$ to state
$X$, and $w_h\geq1$ is the heuristic inflation factor. Setting $w_h=1$
recovers the standard A* weighting, while $w_h>1$ places greater emphasis on
the heuristic estimate during search.

\subsubsection{Obstacle- and Smoothness-Aware Traversal Cost}
\label{subsubsec:traversal_cost} 

\begin{algorithm}[t]
\caption{Obstacle- and Smoothness-Aware Weighted A* Search}
\label{alg:astar}
\begin{algorithmic}[1]

\State \textbf{Input:} occupancy map $I$, obstacle probability map $P$,
start $S$, goal $G$
\State \textbf{Parameters:} heuristic weight $w_h$, obstacle weight $w_o$,
smoothness weight $w_s$

\State $X_S \gets (\varnothing,S)$
\State Initialize priority queue $\mathcal{O}\gets\{X_S\}$
\State Initialize closed set $\mathcal{C}\gets\emptyset$
\State $g(X_S)\gets0$
\State $f(X_S)\gets w_h\,h_0(S,G)$
\State $\textit{Parent}(X_S)\gets\varnothing$
\State Set $g(X)\gets\infty$ for all unvisited states $X$

\While{$\mathcal{O}\neq\emptyset$}

    \State $X\gets
    \operatorname*{arg\,min}_{Y\in\mathcal{O}} f(Y)$
    \State $\mathcal{O}\gets\mathcal{O}\setminus\{X\}$

    \State Extract $(Q_p,Q)$ from $X$

    \If{$Q=G$}
        \State \Return path obtained by backtracking
        $\textit{Parent}$ from $X$ to $X_S$
    \EndIf

    \State $\mathcal{C}\gets\mathcal{C}\cup\{X\}$

    \For{each $Q_n\in\mathcal{N}(Q)$}

        \If{$I(Q_n)=1$}
            \State \textbf{continue}
        \EndIf

        \State $X_n\gets(Q,Q_n)$

        \State $c_{\mathrm{turn}}
        \gets
        c_{\mathrm{smooth}}(Q_p,Q,Q_n)$

        \State $g_{\mathrm{tent}}
        \gets
        g(X)
        +
        d_{\mathrm{step}}(Q,Q_n)
        +
        w_o\,P(Q_n)
        +
        w_s\,c_{\mathrm{turn}}$

        \If{$g_{\mathrm{tent}}<g(X_n)$}

            \State $\textit{Parent}(X_n)\gets X$
            \State $g(X_n)\gets g_{\mathrm{tent}}$

            \State $f(X_n)
            \gets
            g(X_n)
            +
            w_h\,h_0(Q_n,G)$

            \If{$X_n\in\mathcal{C}$}
                \State $\mathcal{C}
                \gets
                \mathcal{C}\setminus\{X_n\}$
            \EndIf

            \State Insert or update $X_n$ in $\mathcal{O}$
            with priority $f(X_n)$

        \EndIf

    \EndFor

\EndWhile

\State \Return ``No valid path found''

\end{algorithmic}
\end{algorithm}

For a transition from the current state $X=(Q_p,Q)$ to a neighboring location
$Q_n$, the successor state is defined as 
\begin{equation}
X_n
=
(Q,Q_n).
\label{eq:successor_state}
\end{equation} 
The basic movement cost between two adjacent locations is 
\begin{equation}
d_{\mathrm{step}}(Q,Q_n)
=
\left\|
Q_n-Q
\right\|_2.
\label{eq:step_cost}
\end{equation} 
Thus, $d_{\mathrm{step}}(Q,Q_n)=1$ for horizontal or vertical moves and,
when diagonal movements are allowed, $\sqrt{2}$ for diagonal moves. 
To discourage abrupt direction changes, we compute a smoothness penalty from
the incoming and outgoing motion vectors. For $Q_p\neq\varnothing$, we define 
\begin{equation}
\mathbf{v}_{\mathrm{in}}
=
Q-Q_p,
\qquad
\mathbf{v}_{\mathrm{out}}
=
Q_n-Q.
\label{eq:motion_vectors}
\end{equation} 
The corresponding turning cost is 
\begin{equation}
c_{\mathrm{smooth}}(Q_p,Q,Q_n)
=
\begin{cases}
0,
&
Q_p=\varnothing,
\\[2mm]
\arccos
\left(
\operatorname{clip}
\left(
\dfrac{
\mathbf{v}_{\mathrm{in}}^{\top}
\mathbf{v}_{\mathrm{out}}
}{
\|\mathbf{v}_{\mathrm{in}}\|_2
\|\mathbf{v}_{\mathrm{out}}\|_2
},
-1,1
\right)
\right),
&
\text{otherwise}.
\end{cases}
\label{eq:smoothness}
\end{equation} 
Here, $\operatorname{clip}(x,-1,1)$ prevents numerical errors caused by
floating-point rounding. Continuing in the same direction yields zero turning
cost, whereas larger changes in direction incur larger penalties. 
We define the incremental traversal cost from $X=(Q_p,Q)$ to
$X_n=(Q,Q_n)$ as 
\begin{equation}
c(X,X_n)
=
d_{\mathrm{step}}(Q,Q_n)
+
w_o\,P(Q_n)
+
w_s\,c_{\mathrm{smooth}}(Q_p,Q,Q_n),
\label{eq:incremental_cost}
\end{equation} 
where $P(Q_n)\in[0,1]$ is the normalized obstacle-related score at $Q_n$,
$w_o\geq0$ controls the contribution of obstacle proximity, and $w_s\geq0$
controls the smoothness penalty. Consequently, locations with stronger
obstacle-related influence incur higher traversal costs, while abrupt turns
are explicitly discouraged. 
Given a candidate transition to $X_n$, the tentative cumulative cost is 
\begin{equation}
g_{\mathrm{tent}}(X_n)
=
g(X)
+
c(X,X_n).
\label{eq:tentative_cost}
\end{equation} 
If $g_{\mathrm{tent}}(X_n)$ is lower than the previously recorded cost for
$X_n$, the cumulative cost and parent state are updated. The priority of the
updated state is subsequently evaluated using
Equation~\ref{eq:weighted_total_cost}.

Algorithm~\ref{alg:astar} initializes the search from
$X_S=(\varnothing,S)$ and maintains an open priority queue
$\mathcal{O}$ ordered by the weighted evaluation function in
Equation~\ref{eq:weighted_total_cost}. At each iteration, the state with the
lowest priority is selected for expansion. If its current location is the goal
$G$, the final path is recovered by backtracking through the stored parent
states. For each traversable neighbor $Q_n\in\mathcal{N}(Q)$, the algorithm constructs
a successor state $X_n=(Q,Q_n)$ and evaluates its tentative cumulative cost
according to Equation~\ref{eq:tentative_cost}. The cost jointly accounts for
movement distance, the learned obstacle-related score, and the turning angle
induced by the candidate transition. If the tentative cost improves upon the
best cost previously recorded for the same state, its cumulative cost, parent,
and priority are updated accordingly. A previously expanded state is reopened
when a lower-cost path to that state is discovered. 
Unlike approaches that smooth a discrete path only after search, our
formulation incorporates the turning penalty directly into the transition
cost. Consequently, obstacle clearance and path smoothness influence the
search itself rather than being treated solely as post-processing objectives.
\section{RESULT}\label{sec:result}

\subsection{Experimental Setup}
\label{subsec:experimental_setup}

\subsubsection{Platform}
\label{subsubsec:platform}

We conduct all experiments on a Linux server equipped with a
16-vCPU Intel(R) Xeon(R) Platinum 8352V CPU and two NVIDIA RTX 4090
GPUs, each with 24\,GB of memory.

\subsubsection{Dataset}
\label{subsubsec:dataset}

We evaluate $\Design$ on three public pathfinding datasets following
the experimental settings commonly adopted in prior learning-based
pathfinding studies~\cite{yonetani2021path,kirilenko2023transpath}:

\begin{itemize}

    \item \textbf{Motion Planning (MP) Dataset}~\cite{bhardwaj2017learning}:
    The MP dataset contains 1,000 environment maps with diverse and
    challenging obstacle configurations, including bug traps and narrow
    gaps. We apply rotation and translation augmentations to obtain
    36,000 maps of size $64\times64$.

    \item \textbf{Bugtrap Forest (BF) Dataset}~\cite{bhardwaj2017learning}:
    The BF dataset contains 1,000 samples covering eight types of
    grid-world environments with distinct obstacle layouts. We augment
    the maps through mirroring and rotation, resulting in 64,000 maps
    of size $64\times64$.

    \item \textbf{Tiled Motion Planning (TMP) Dataset}
    ~\cite{kirilenko2023transpath}:
    TMP is a modified version of the MP dataset in which each map is
    constructed from four MP maps, yielding 4,000 original maps of size
    $64\times64$. We further apply mirroring and rotation to obtain
    64,000 maps.

\end{itemize} 
The proposed obstacle probability map construction module is trained
directly from the environment maps and does not require
heuristic-generated or manually annotated path labels.

\subsection{Model Specification}
\label{subsubsec:model_specification}

We implement $\Design$ using PyTorch. During training, we randomly mask
15\%--20\% of each occupancy map and train the model to reconstruct the
masked regions using the objectives described in
Section~\ref{subsec:probability_map}. 
The Vision Kansformer encoder consists of 12 layers with 6 attention
heads and an embedding dimension of 384. A linearly increasing drop-path
rate from 0 to 0.1 is applied across the encoder layers. 
The reconstruction head contains two MLP layers with an input dimension
of 768, a hidden dimension of 3072, and an output dimension of 768.
Each MLP layer uses a GELU activation with a dropout rate of 0.0. The
KAN module in the reconstruction head has an input and output dimension
of 768, a grid size of 5, a spline order of 3, and SiLU as the base
activation function. 
The reconstruction features are subsequently processed by a transposed
convolution with a kernel size of $16\times16$ and a stride of 16,
producing a three-channel spatial representation. This is followed by
three convolutional layers with $3\times3$ kernels and output channel
dimensions of 64, 32, and 3, respectively. 
We train each model with a batch size of 32 for up to 500 epochs or
until convergence. The initial learning rate is set to $1\times10^{-4}$,
and the model is optimized using AdamW~\cite{loshchilov2017decoupled}.
The learning rate and weight decay follow cosine schedules during
training. Mixed-precision training is employed, and training is
distributed across two NVIDIA RTX 4090 GPUs. We use a random seed of
39 for data shuffling and parameter initialization. 
For contrastive supervision, we use an ImageNet-pretrained Vision
Transformer (ViT)~\cite{dosovitskiy2020image,deng2009imagenet} as the
teacher network, as described in
Section~\ref{subsubsec:training_objectives}. The temperature parameter
$\tau$ in the contrastive loss is set to 0.2. The teacher network is
used only during training to provide reference representations and is
not part of the inference pipeline. 
The weighting coefficients for the reconstruction, perceptual, and
contrastive losses are set to
$\lambda=1.0$, $\alpha=0.1$, and $\beta=0.1$, respectively. 
For the obstacle- and smoothness-aware path search, we use a
4-connected neighborhood. The heuristic weight is set to
$w_h=1.0$, the obstacle-proximity weight to $w_o=0.5$, and the
smoothness weight to $w_s=0.2$. These parameters correspond to the
weighted evaluation and traversal costs defined in
Section~\ref{subsec:weighted_astar}.

\subsection{Baselines}
\label{subsubsec:baselines}

We compare $\Design$ with the following representative search-based and
learning-based methods:
\begin{itemize}
    \item \textbf{Weighted A* Search (WA*)}~\cite{pohl1970heuristic}:
    WA* extends A* by multiplying the heuristic term by a weighting
    factor. Its evaluation function is
    $f(n)=g(n)+w\,h(n)$, where $w$ controls the relative emphasis on
    heuristic guidance.
    \item \textbf{Path Safety Inflation Planning Algorithm Based on
    Improved JPS (Improved JPS)}~\cite{lin2024path}:
    Improved JPS augments Jump Point Search with a path-safety inflation
    mechanism that discourages traversal near obstacles while retaining
    the search-space reduction provided by JPS.
    \item \textbf{Self-Adaptive Improved Learning Search (SAIL-SL)}
    ~\cite{choudhury2018data}:
    SAIL-SL employs learning-based search guidance to improve planning
    efficiency by adapting its search behavior to characteristics of
    the environment.
    \item \textbf{Branch-and-Bound A* (BB-A*)}
    ~\cite{vlastelica2019differentiation}:
    BB-A* combines A*-based search with branch-and-bound operations to
    reduce unnecessary exploration while solving shortest-path
    problems. 
    \item \textbf{Five-Neighborhood Dynamic Weighted A* (FDW-A*)}
    ~\cite{han2023mobile}:
    FDW-A* incorporates a five-neighborhood expansion strategy and
    dynamically adjusts heuristic weighting to balance search
    efficiency and path quality.
    \item \textbf{Artificial Potential Field Enhanced A* (APF-A*)}
    ~\cite{zhang2024agv}:
    APF-A* integrates artificial potential field information into A*
    search, using obstacle-induced repulsive effects to encourage paths
    with improved obstacle clearance.
\end{itemize}

\subsection{Evaluation Metrics}
\label{subsubsec:evaluation_metrics}

We evaluate each method in terms of path safety, obstacle clearance,
trajectory smoothness, obstacle-related traversal cost, search
efficiency, and latency.

\begin{itemize}
    \item \textbf{Path Safety Score:}
    The path safety score measures the accumulated clearance between
    locations on the generated path and their nearest obstacles. For a
    path containing $N$ locations, it is defined as
    \begin{equation}
    S_{\mathrm{safety}}
    =
    \sum_{i=1}^{N}
    d
    \left(
    Q_i,
    \mathcal{O}_{\mathrm{nearest}}(Q_i)
    \right),
    \label{eq:path_safety_score}
    \end{equation}
    where
    $d(Q_i,\mathcal{O}_{\mathrm{nearest}}(Q_i))$
    denotes the Euclidean distance from path location $Q_i$ to its
    nearest obstacle. A higher score indicates greater accumulated
    obstacle clearance along the path.

    \item \textbf{Average Obstacle Distance:}
    To reduce the dependence of the safety score on path length, we
    additionally report the mean distance from path locations to their
    nearest obstacles:
    \begin{equation}
    D_{\mathrm{obs}}
    =
    \frac{1}{N}
    \sum_{i=1}^{N}
    d
    \left(
    Q_i,
    \mathcal{O}_{\mathrm{nearest}}(Q_i)
    \right).
    \label{eq:average_obstacle_distance}
    \end{equation}
    A higher value indicates greater average obstacle clearance.

    \item \textbf{Angular Deviation Score:}
    We quantify directional consistency using the standard deviation of
    the orientations of consecutive path segments. Let $\theta_i$
    denote the orientation of the segment from $Q_i$ to $Q_{i+1}$.
    The angular deviation score is 
    \begin{equation}
    S_{\mathrm{angle}}
    =
    \operatorname{std}
    \left(
    \theta_1,\theta_2,\ldots,\theta_{N-1}
    \right).
    \label{eq:angular_deviation}
    \end{equation}
    A lower angular deviation score indicates smaller variation in
    segment orientations and therefore greater directional consistency
    along the generated path.

    \item \textbf{Average Obstacle-Related Cost:}
    Because the learned probability map $P$ represents a normalized
    obstacle-related score, we measure the average value encountered
    along a generated path as
    \begin{equation}
    C_{\mathrm{obs}}
    =
    \frac{1}{N}
    \sum_{i=1}^{N}
    P(Q_i).
    \label{eq:average_obstacle_cost}
    \end{equation} 
    A lower value indicates that the generated path traverses regions
    with weaker obstacle-related influence.

    \item \textbf{Node Exploration Reduction:}
    We measure search efficiency by comparing the number of explored
    states with that of standard A*. Let $E_{\mathrm{A^*}}$ denote the
    number of states explored by A* and $E$ the number explored by the
    evaluated method. Following our evaluation protocol, the reduction
    is computed as 
    \begin{equation}
    R_{\mathrm{explore}}
    =
    \max
    \left(
    100
    \times
    \frac{
    E_{\mathrm{A^*}}-E
    }{
    E_{\mathrm{A^*}}
    },
    0
    \right).
    \label{eq:node_reduction}
    \end{equation} 
    A higher value indicates that fewer states are explored relative to
    standard A*.

    \item \textbf{Inference Latency:}
    Inference latency measures the elapsed time from the start of the
    path search to retrieval of the final path: 
    \begin{equation}
    T_{\mathrm{inf}}
    =
    T_{\mathrm{end}}
    -
    T_{\mathrm{start}},
    \label{eq:inference_latency}
    \end{equation} 
    where $T_{\mathrm{start}}$ and $T_{\mathrm{end}}$ denote the
    timestamps at the beginning and end of the path search,
    respectively. Lower latency indicates more efficient path search.

\end{itemize}

\subsection{Qualitative results}

\begin{figure*}[t] 
\centering
    \includegraphics[width=\textwidth]{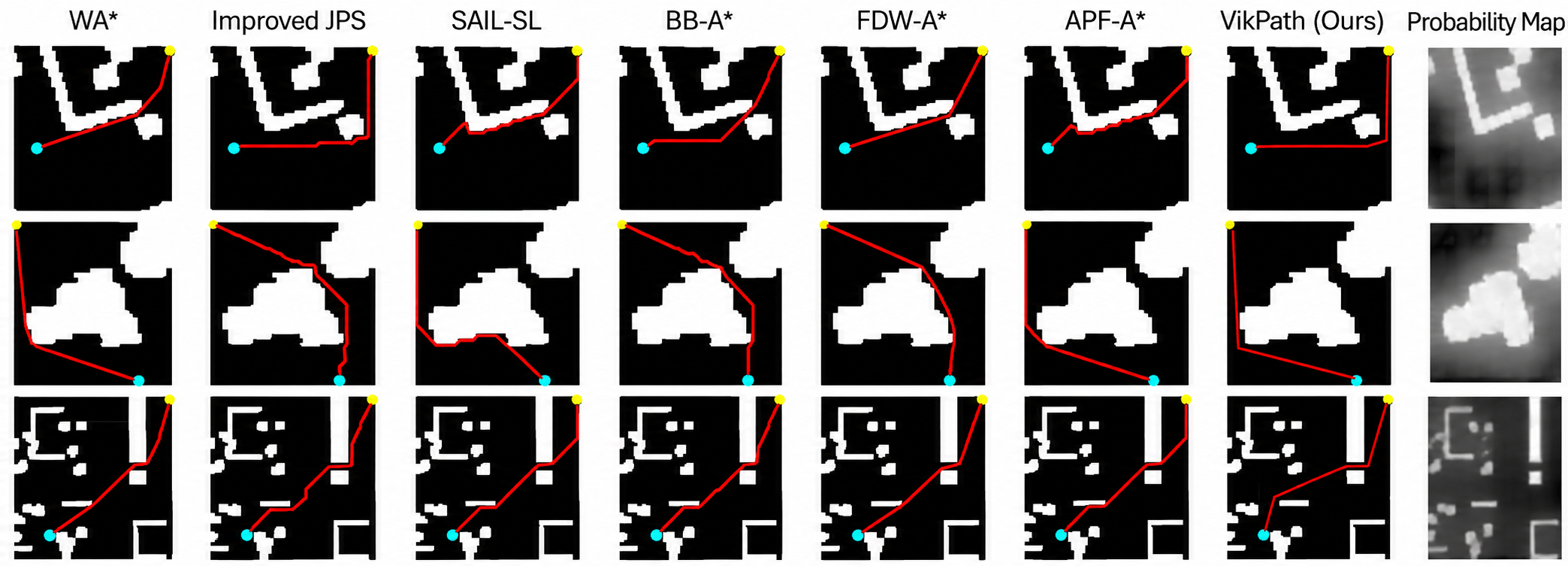}
    \caption{\textbf{Selected Pathfinding Results}: White pixels indicate obstacles. Starting points are marked in cyan, destinations are marked in yellow, and the generated paths are annotated in red. The rightmost column is our probability map. }
    \label{fig:results1}
\end{figure*}


Figure~\ref{fig:results1} provides qualitative examples of the paths generated
by different methods, together with the obstacle probability map produced by
$\Design$.
In the probability map, white regions correspond to obstacles and are
non-traversable, while lighter gray regions indicate higher obstacle-related
scores and stronger proximity to nearby obstacles. Darker gray regions
correspond to lower-cost areas with greater obstacle clearance, and black
regions represent relatively open free space. 
Compared with the baseline methods, $\Design$ more effectively avoids regions
with strong obstacle-related influence, resulting in paths with greater
clearance from nearby obstacles. By incorporating the learned probability map
into the search objective, $\Design$ exploits global spatial information about
the environment before and during path search, allowing the planner to favor
safer regions instead of relying solely on local geometric distance to the
goal. Moreover, the explicit smoothness penalty discourages abrupt direction
changes, enabling $\Design$ to balance obstacle clearance, path efficiency,
and trajectory smoothness.

\subsection{Quantitative results}
\begin{table*}[ht]
\footnotesize
\caption{Comparison of VikPath with baselines. The best (second-best) results are highlighted in bold (underlined).}
\label{tab:result1}
\begin{adjustbox}{width=\textwidth,center}
    \centering
    \begin{tabular}{c| c | c c c c c c c} 
    \toprule 
    Dataset & Evaluation Metrics 
    & WA*~\cite{pohl1970heuristic} & Improved JPS~\cite{lin2024path} 
    & SAIL-SL~\cite{choudhury2018data} & BB-A*~\cite{vlastelica2019differentiation} & FDW-A*~\cite{han2023mobile} & APF-A*~\cite{zhang2024agv}  
    & VikPath (Ours) \\ \midrule

    \multirow{6}{*}{MP}  
    & Path Safety Score  & 6.9876 & \underline{8.6894}  & 6.7856 & 7.2311 & 6.4317 & 7.0412 & $\textbf{8.7925}$ \\  
    & Average Obstacle Distance & 684.3229 & $\underline{773.5716}$  & 598.1166 & 698.1166 & 655.6904 & 703.4987 & $\textbf{842.2555}$ \\
    & Angular Deviation Score & 0.3388 & 0.3369  & 0.3073 & 0.3376 & \textbf{0.1958} & 0.3068 & \underline{0.2382} \\ 
    & Average Path Cost & \underline{0.1890} & 0.1902  & 0.1974 & 0.1898 & 0.2183 & 0.1992 & \textbf{0.1844} \\ 
    & Node Exploration Reduction & 58.60 & 24.30  & $\underline{71.29}$ & 52.74 & 64.58 & 18.90 & \textbf{77.19} \\ 
    & Inference Latency & 0.5905 & 0.1687  & 5.1634 & \underline{0.1510} & 0.3653 & 5.8769 & \textbf{0.0169} \\  
    \midrule
    
    \multirow{6}{*}{BF} 
    & Path Safety Score  & 6.3569 & \underline{7.4206}  & 6.3538 & 6.2080 & 6.0156 & 6.5630 & $\textbf{7.5218}$ \\  
    & Average Obstacle Distance & 577.8936 & $\underline{656.6303}$  & 573.7912 & 654.1192 & 554.7262 & 648.3308 & $\textbf{680.9704}$ \\
    & Angular Deviation Score & 0.3011 & 0.3135  & 0.2779 & 0.2992 & \textbf{0.1144} & 0.3086 & \underline{0.2587} \\ 
    & Average Path Cost & 0.0906 & \underline{0.0801}  & 0.1021 & 0.0876 & 0.1101 & 0.0929 & \textbf{0.0726} \\ 
    & Node Exploration Reduction & 65.02 & 22.00  & \textbf{77.67} & 58.74 & 65.45 & 17.40 & \underline{71.18} \\ 
    & Inference Latency & 0.2545 & 0.2027  & 2.7959 & \underline{0.1112} & 0.1852 & 6.0321 & \textbf{0.0155} \\  
    \midrule

    \multirow{6}{*}{TMP} 
    & Path Safety Score  & 5.2545 & \underline{6.1630}  & 5.2791 & 5.3901 & 4.7082 & 5.2002 & $\textbf{6.6140}$ \\  
    & Average Obstacle Distance & 575.0037 & $\underline{605.1801}$  & 554.7428 & 539.4223 & 455.5975 & 565.2684 & $\textbf{631.1702}$ \\
    & Angular Deviation Score & 0.3064 & 0.3210  & 0.2971 & 0.3185 & \textbf{0.1462} & 0.3285 & \underline{0.2682} \\ 
    & Average Path Cost & 0.1922 & \underline{0.1864}  & 0.1956 & 0.1899 & 0.2215 & 0.1941 & \textbf{0.1855} \\ 
    & Node Exploration Reduction & 59.69 & 19.97  & \textbf{82.97} & 56.64 & 56.58 & 19.64 & \underline{71.52} \\ 
    & Inference Latency & 0.3436 & 0.1580  & 2.5971 & \underline{0.1488} & 0.3281 & 6.7392 & \textbf{0.0192} \\  
    \bottomrule
    \end{tabular}
\end{adjustbox}
\end{table*}

\subsubsection{Model Performance}

As detailed in Table~\ref{tab:result1}, $\Design$ consistently achieves
higher path safety scores and average obstacle distances than the baseline
methods, demonstrating its ability to generate paths with greater obstacle
clearance. $\Design$ also obtains a low angular deviation score and a low
average obstacle-related cost, indicating that the generated paths avoid
high-risk regions while maintaining smooth trajectories. 
Specifically, compared with the second-best method, $\Design$ achieves, on
average, a 3.28\% higher path safety score, a 5.62\% higher average obstacle
distance, and a 4.09\% lower average obstacle-related cost. Although
$\text{FDW-A}^*$ achieves the lowest angular deviation score and therefore
produces the smoothest paths according to this metric, it performs worse on
the safety-related metrics. On average, $\text{FDW-A}^*$ exhibits a 34.4\%
lower path safety score, a 29.91\% lower average obstacle distance, and a
29.38\% higher average obstacle-related cost than $\Design$. 
These results reveal an important trade-off between trajectory smoothness and
obstacle clearance. Post-processing a path for additional smoothness can reduce
angular variation, but may also move portions of the trajectory closer to
nearby obstacles. In contrast, $\Design$ incorporates the turning penalty and
the learned obstacle-related cost directly into the search objective. As a
result, it achieves the second-best angular deviation score while maintaining
substantially greater obstacle clearance and lower obstacle-related cost. This
demonstrates that $\Design$ provides a favorable balance among path safety,
efficiency, and trajectory smoothness.

\subsubsection{Efficiency}
As shown in Table~\ref{tab:result1}, $\Design$ achieves the highest node
exploration reduction on MP and ranks second on BF and TMP. These results show
that $\Design$ can reduce the number of states explored during search relative
to standard A* and the evaluated pathfinding baselines. Its slightly lower
ranking on BF and TMP reflects the trade-off introduced by explicitly
considering obstacle clearance and trajectory smoothness in addition to
goal-directed search efficiency. 
$\Design$ also achieves the lowest inference latency across all three datasets,
outperforming the second-best baseline by an average of 87.06\%. The learned
obstacle probability map provides global spatial information about obstacle
proximity before path search begins, allowing the planner to assign higher
costs to less desirable regions and prioritize safer search directions. This
reduces unnecessary exploration while retaining the ability to generate
high-quality paths. 
Overall, the results demonstrate that $\Design$ combines low inference latency
and efficient search with strong obstacle clearance and trajectory smoothness,
making it well suited for latency-sensitive pathfinding applications in
complex environments.

\begin{table*}[ht]
\caption{Evaluation of model generalization. The best (second-best) results are highlighted in bold (underlined).}
\label{tab:result_generalizability}
\footnotesize
\begin{adjustbox}{width=\textwidth,center}
    \centering
    \begin{tabular}{c| c | c c c c c c c} 
    \toprule 
    Dataset & Evaluation Metrics 
    & WA*~\cite{pohl1970heuristic} & Improved JPS~\cite{lin2024path} 
    & SAIL-SL~\cite{choudhury2018data} & BB-A*~\cite{vlastelica2019differentiation} & FDW-A*~\cite{han2023mobile} & APF-A*~\cite{zhang2024agv}  
    & VikPath (Ours) \\ \midrule

    \multirow{6}{*}{\begin{tabular}[c]{@{}c@{}}TMP\\ \\$\downarrow$ \\ \\City\end{tabular}}  
    & Path Safety Score  & 3.3263 & \underline{3.7047}  & 3.3354 & 3.3941 & 3.6025 & 3.6325 & $\textbf{3.8100}$ \\  
    & Average Obstacle Distance & 70.4603 & $\underline{105.3754}$  & 98.7047 & 104.8952 & 63.4175 & 76.4469 & $\textbf{117.5299}$ \\
    & Angular Deviation Score & 0.3445 & 0.3974  & 0.3740 & 0.3482 & \textbf{0.1680} & 0.3571 & \underline{0.2103} \\ 
    & Average Path Cost & 0.1756 & 0.2149 & 0.1790 & 0.1726 & 0.1896 & \textbf{0.1678} & \textbf{0.1635} \\ 
    & Node Exploration Reduction & 70.97 & 24.26  & \textbf{83.30} & 60.57 & 65.55 & 23.08 & \underline{72.18} \\ 
    & Inference Latency & 0.2801 & 0.1419  & 3.8586 & \underline{0.1360} & 0.3074 & 8.2751 & \textbf{0.0179} \\ 
    \midrule

    \multirow{6}{*}{\begin{tabular}[c]{@{}c@{}}TMP\\ \\$\downarrow$ \\ \\Game\end{tabular}} 
    & Path Safety Score  & 2.6901 & 2.8809 & 2.6299 & 2.7677 & \underline{3.0186} & 2.8038 & $\textbf{3.1918}$ \\   
    & Average Obstacle Distance & 192.4229 & $\underline{290.2455}$  & 181.4689 & 279.9509 & 170.3953 & 192.0487 & $\textbf{362.8154}$ \\
    & Angular Deviation Score & 0.3754 & 0.3883  & 0.3940 & 0.3785 & \textbf{0.2202} & 0.3764 & \underline{0.3696} \\ 
    & Average Path Cost & 0.2469 & 0.2671  & 0.2578 & \underline{0.2433} & 0.2769 & 0.2499 & \textbf{0.2399} \\ 
    & Node Exploration Reduction & 52.86 & 22.97  & \underline{57.72} & 30.42 & 37.58 & 12.56 & \textbf{61.99} \\ 
    & Inference Latency & 0.3523 & \underline{0.0752}  & 5.0898 & 0.0986 & 0.2436 & 4.9188 & \textbf{0.0098} \\
    \bottomrule
    \end{tabular}
\end{adjustbox}

\end{table*}

\subsubsection{Geralization}

To evaluate the generalization capability of $\Design$, we directly apply the
model trained on the TMP dataset to two unseen datasets, City and
Game~\cite{kirilenko2023transpath}, without any additional fine-tuning.
The City dataset contains 1,000 maps of size $64\times64$ derived from urban
street scenes, while the Game dataset contains 1,000 maps of size
$64\times64$ derived from game environments. These datasets exhibit obstacle
layouts and spatial structures that differ from those observed during
training, providing a challenging setting for evaluating cross-dataset
generalization. 
As shown in Table~\ref{tab:result_generalizability}, $\Design$ consistently
performs favorably on both unseen datasets. On the City dataset, compared with
the second-best method, $\Design$ achieves, on average, a 2.84\% higher path
safety score, a 25.00\% higher average obstacle distance, and a 2.62\% lower
average obstacle-related cost. $\Design$ also achieves the lowest inference
latency, outperforming the second-best method by 86.83\% on average. 
On the Game dataset, $\Design$ achieves, on average, a 5.73\% higher path
safety score, an 11.53\% higher average obstacle distance, a 1.41\% lower
average obstacle-related cost, and a 7.39\% higher node exploration reduction
than the second-best approach. It also achieves the lowest inference latency,
with an average reduction of 86.96\% relative to the second-best method. 
These results demonstrate that the obstacle-aware representations learned by
$\Design$ transfer effectively to previously unseen environments without
fine-tuning. In particular, $\Design$ maintains strong obstacle clearance and
low obstacle-related traversal cost while preserving high search efficiency,
indicating robust cross-dataset generalization across environments with
substantially different spatial structures.

\vspace{-1mm}
\subsection{Ablation Study}
\label{subsec:ablation}

\begin{table*}[]
\caption{Comparison of Pathfinding Cost Function}
\label{tab:ablation2}
\footnotesize
\begin{adjustbox}{width=\textwidth,center}
\centering
\begin{tabular}{c| c | c c c c c c c}
\toprule
Dataset&Cost Function & \begin{tabular}[c]{@{}c@{}} Path Safety \\ Score\end{tabular} &   \begin{tabular}[c]{@{}c@{}}Average  \\ Obstacle Distance\end{tabular} &   \begin{tabular}[c]{@{}c@{}}Angular \\ Deviation Score \end{tabular} &  \begin{tabular}[c]{@{}c@{}}Average \\ Path Cost \end{tabular}   & \begin{tabular}[c]{@{}c@{}}Node Exploration \\Reduction\end{tabular}
 & \begin{tabular}[c]{@{}c@{}}Inference \\Latency\end{tabular}\\ \midrule
\multirow{4}{*}{\text{MP}} &Improved A* (w/ obstacle proximity) & $\underline{8.7264}$ & $\underline{824.6473}$ & $0.4685$ & $0.2417$ & $9.78$ & $0.9562$ \\
&VikPath (w/o obstacle proximity) & $6.7417$ & $564.9778$ & $\underline{0.2875}$ & $0.2074$ & $57.60$ & $\underline{0.0505}$ \\
&VikPath (w/o Smoothing) &  $8.0547$ & $762.4502$ & $0.5361$ & $\underline{0.1866}$ & $\underline{60.68}$ & $0.0518$ \\
& VikPath  & $\textbf{8.7925}$ & $\textbf{842.2555}$ & $\textbf{0.2382}$ & $\textbf{0.1844}$ & $\textbf{77.19}$ & $\textbf{0.0169}$ \\ \midrule

\multirow{4}{*}{\text{BF}} &Improved A* (w/ obstacle proximity) & $\underline{7.5096}$ & $619.1945$ & $0.3152$ & $0.2232$ & $16.53$ & $1.1477$ \\
&VikPath (w/o obstacle proximity) & $6.1504$ & $554.1531$ & $\underline{0.2757}$ & $0.1877$ & $62.38$ & $\textbf{0.0039}$ \\
&VikPath (w/o Smoothing) & $7.3085$ & $\underline{650.8057}$ & $0.5834$ & $\underline{0.0841}$ & $\underline{70.93}$ & $0.0300$ \\ 
&VikPath  & $\textbf{7.5218}$ & $\textbf{680.9704}$ & $\textbf{0.2587}$ & $\textbf{0.0726}$ & $\textbf{71.18}$ & $\underline{0.0155}$ \\ \midrule

\multirow{4}{*}{\text{TMP}} &Improved A* (w/ obstacle proximity) & $\underline{6.5419}$ & $582.2399$ & $0.3106$ & $0.3016$ & $29.56$ & $0.9186$ \\
&VikPath (w/o obstacle proximity) & $5.6355$ & $554.1340$ & $\underline{0.2897}$ & $0.2705$ & $\underline{68.30}$ & $\underline{0.0589}$ \\
&VikPath (w/o Smoothing) & $6.3176$ & $\underline{593.7170}$ & $0.5315$ & $\underline{0.2190}$ & $60.35$ & $0.0604$ \\
&VikPath & $\textbf{6.6140}$ & $\textbf{631.1702}$ & $\textbf{0.2682}$ & $\textbf{0.1875}$ & $\textbf{71.52}$ & $\textbf{0.0192}$ \\
\bottomrule
\end{tabular}
\end{adjustbox}
\end{table*}

\subsubsection{Comparison of Cost Functions}
\label{subsubsec:cost_ablation}

To evaluate the contribution of individual components in the traversal cost
defined in Equation~\ref{eq:incremental_cost}, we compare the full $\Design$
with three variants:
(\romannumeral 1) an improved A* method that incorporates obstacle proximity
using geometrically computed distances to nearby obstacles;
(\romannumeral 2) $\Design$ without the obstacle-related penalty derived from
the learned probability map $P$; and
(\romannumeral 3) $\Design$ without the smoothness penalty for discouraging
abrupt direction changes. 
As shown in Table~\ref{tab:ablation2}, jointly incorporating the learned
obstacle-related cost and the smoothness penalty provides the most favorable
overall performance across path safety, obstacle clearance, smoothness, and
search efficiency. 
Compared with the variant without the learned obstacle-related penalty,
$\Design$ achieves, on average, a 23.35\% higher path safety score, a 28.61\%
higher average obstacle distance, an 11.73\% lower angular deviation score,
an 11.27\% lower average obstacle-related cost, and a 17.60\% higher node
exploration reduction. These results demonstrate that the learned obstacle
probability map provides useful spatial guidance beyond geometric
goal-directed search alone. 
Compared with the variant without the smoothness penalty, $\Design$ achieves,
on average, a 5.58\% higher path safety score, a 7.13\% higher average
obstacle distance, a 53.85\% lower angular deviation score, a 9.74\% lower
average obstacle-related cost, and a 15.35\% higher node exploration
reduction. The substantial reduction in angular deviation confirms the
importance of explicitly incorporating directional changes into the
traversal cost. 
Compared with the improved A* variant that incorporates geometrically computed
obstacle proximity, $\Design$ achieves a 0.67\% higher path safety score,
a 6.83\% higher average obstacle distance, a 26.90\% lower angular deviation
score, a 43.00\% lower average obstacle-related cost, and a 74.25\% higher
node exploration reduction. In addition, $\Design$ achieves, on average,
97.34\% lower inference latency than the improved A* baseline. 
Overall, these results show that the learned obstacle probability map and the
explicit smoothness penalty play complementary roles. The former provides
environment-dependent obstacle-aware guidance, while the latter discourages
abrupt turns during search. Their combination enables $\Design$ to achieve a
strong balance among obstacle clearance, trajectory smoothness, and search
efficiency.

\begin{table}[ht]
  \caption{Comparison of Image Reconstruction Techniques}
    \label{tab:ablation1}
\begin{adjustbox}{width=0.6\columnwidth,center}
    \centering
\footnotesize
\begin{tabular}{ c|c c c c} \toprule 
Dataset & Model & MSE & PSNR & SSIM \\ \midrule
\multirow{3}{*}{\text{MP}} 
& Vision Transformer & $0.0534$ & $ 22.7308 $ & $ 0.7052$ \\
& Vision Mamba &  $ 0.1269 $ & $ 12.7986 $ & $ 0.6241$ \\
& Vision Kansformer (Ours) &$\textbf{0.0325} $ & $ \textbf{24.8953}$ & $ \textbf{0.7893}$\\ \midrule
\multirow{3}{*}{\text{BF}} 
& Vision Transformer  & $0.0583 $ & $22.3611 $ & $ 0.6896$\\
& Vision Mamba & $0.1365 $ &$14.3792$ & $0.6473$ \\
& Vision Kansformer (Ours) &  $\textbf{0.0292}$ & $\textbf{25.3793}$ &$\textbf{0.8074}$  \\ \midrule
\multirow{3}{*}{\text{TMP}} 
& Vision Transformer &$0.0216 $ & $26.6776 $ & $ 0.8425$\\
& Vision Mamba & $0.0817 $ & $15.9879 $ & $0.7042 $\\
& Vision Kansformer (Ours) &$\textbf{0.0123}$ &$\textbf{29.1451}$&$\textbf{0.9066}$\\ 
    \bottomrule
  \end{tabular}
 \end{adjustbox}
 \vspace{-1mm}
\end{table}

\subsubsection{Comparison of Image Reconstruction Techniques}
\label{subsubsec:reconstruction_ablation}

To evaluate the effectiveness of Vision Kansformer for learning
obstacle-aware representations through masked reconstruction, we compare it
with (\romannumeral 1) SiT~\cite{ahmed2021sit}, a Transformer-based
self-supervised vision model, and (\romannumeral 2) Vision
Mamba~\cite{zhu2024vision}, a state-space-based vision backbone. 
We evaluate reconstruction quality using mean squared error (MSE), peak
signal-to-noise ratio (PSNR), and structural similarity index measure
(SSIM)~\cite{sara2019image}. Lower MSE and higher PSNR and SSIM indicate
better reconstruction quality. 
As shown in Table~\ref{tab:ablation1}, Vision Kansformer achieves consistently
better reconstruction performance across the evaluated datasets than SiT and
Vision Mamba. These results suggest that combining self-attention with
KAN-based nonlinear transformations is effective for modeling the spatial
structure of occupancy maps. Self-attention captures long-range dependencies
among spatially separated regions, while the spline-based transformations in
KAN provide flexible nonlinear modeling of local feature relationships.
Together, these components allow Vision Kansformer to preserve both global
environmental structure and fine-grained obstacle-related information during
reconstruction.
\section{Conclusions}
\label{sec:conclusion}

In this paper, we present $\Design$, a self-supervised pathfinding framework
that integrates learned obstacle-aware representations with
smoothness-aware search. By learning an obstacle probability map through
Vision Kansformer and incorporating obstacle proximity and turning penalties
directly into the search objective, $\Design$ generates paths that achieve a
favorable balance among obstacle clearance, trajectory smoothness, and search
efficiency. Extensive experiments demonstrate consistent improvements over
existing pathfinding baselines, together with low inference latency and strong
generalization to unseen environments. These results highlight the potential
of $\Design$ as an efficient and generalizable framework for pathfinding in
complex obstacle-rich environments.
\newpage

\bibliographystyle{ACM-Reference-Format}
\bibliography{refenrece}

@inproceedings{lee2018gated,
  title={Gated Path Planning Networks},
  author={Lee, Alexander and Popovic, Jovan and Koltun, Vladlen},
  booktitle={Proceedings of the 35th International Conference on Machine Learning (ICML)},
  year={2018}
}

@article{panov2018grid,
  title={Grid path planning with deep reinforcement learning: Preliminary results},
  author={Panov, Aleksandr I and Yakovlev, Konstantin S and Suvorov, Roman},
  journal={Procedia computer science},
  volume={123},
  pages={347--353},
  year={2018},
  publisher={Elsevier}
}

@inproceedings{kirilenko2023transpath,
  title={Transpath: Learning heuristics for grid-based pathfinding via transformers},
  author={Kirilenko, Daniil and Andreychuk, Anton and Panov, Aleksandr and Yakovlev, Konstantin},
  booktitle={Proceedings of the AAAI Conference on Artificial Intelligence},
  volume={37},
  number={10},
  year={2023}
}

@article{hart1968formal,
  title={A formal basis for the heuristic determination of minimum cost paths},
  author={Hart, Peter E and Nilsson, Nils J and Raphael, Bertram},
  journal={IEEE transactions on Systems Science and Cybernetics},
  volume={4},
  number={2},
  pages={100--107},
  year={1968},
  publisher={IEEE}
}

@incollection{dijkstra2022note,
  title={A note on two problems in connexion with graphs},
  author={Dijkstra, Edsger W},
  booktitle={Edsger Wybe Dijkstra: His Life, Work, and Legacy},
  pages={287--290},
  year={2022}
}

@article{dosovitskiy2020image,
  title={An image is worth 16x16 words: Transformers for image recognition at scale},
  author={Dosovitskiy, Alexey and Beyer, Lucas and Kolesnikov, Alexander and Weissenborn, Dirk and Zhai, Xiaohua and Unterthiner, Thomas and Dehghani, Mostafa and Minderer, Matthias and Heigold, Georg and Gelly, Sylvain and others},
  journal={arXiv preprint arXiv:2010.11929},
  year={2020}
}

@article{liu2024kan,
  title={Kan: Kolmogorov-arnold networks},
  author={Liu, Ziming and Wang, Yixuan and Vaidya, Sachin and Ruehle, Fabian and Halverson, James and Solja{\v{c}}i{\'c}, Marin and Hou, Thomas Y and Tegmark, Max},
  journal={arXiv preprint arXiv:2404.19756},
  year={2024}
}

@article{liu2024deep,
  title={Deep reinforcement learning for mobile robot path planning},
  author={Liu, Hao and Shen, Yi and Yu, Shuangjiang and Gao, Zijun and Wu, Tong},
  journal={arXiv preprint arXiv:2404.06974},
  year={2024}
}

@article{karur2021survey,
  title={A survey of path planning algorithms for mobile robots},
  author={Karur, Karthik and Sharma, Nitin and Dharmatti, Chinmay and Siegel, Joshua E},
  journal={Vehicles},
  volume={3},
  number={3},
  pages={448--468},
  year={2021},
  publisher={MDPI}
}

@article{qin2023review,
  title={Review of autonomous path planning algorithms for mobile robots},
  author={Qin, Hongwei and Shao, Shiliang and Wang, Ting and Yu, Xiaotian and Jiang, Yi and Cao, Zonghan},
  journal={Drones},
  volume={7},
  number={3},
  pages={211},
  year={2023},
  publisher={MDPI}
}

@inproceedings{takahashi2019learning,
  title={Learning heuristic functions for mobile robot path planning using deep neural networks},
  author={Takahashi, Takeshi and Sun, He and Tian, Dong and Wang, Yebin},
  booktitle={Proceedings of the International Conference on Automated Planning and Scheduling},
  volume={29},
  pages={764--772},
  year={2019}
}

@inproceedings{yonetani2021path,
  title={Path planning using neural a* search},
  author={Yonetani, Ryo and Taniai, Tatsunori and Barekatain, Mohammadamin and Nishimura, Mai and Kanezaki, Asako},
  booktitle={International conference on machine learning},
  pages={12029--12039},
  year={2021},
  organization={PMLR}
}

@inproceedings{hu2021path,
  title={Path planning with q-learning},
  author={Hu, Yuepeng and Yang, Lehan and Lou, Yizhu},
  booktitle={Journal of Physics: Conference Series},
  volume={1948},
  number={1},
  pages={012038},
  year={2021},
  organization={IOP Publishing}
}

@article{aberdeen2005policy,
  title={Policy-gradient methods for planning},
  author={Aberdeen, Douglas},
  journal={Advances in Neural Information Processing Systems},
  volume={18},
  year={2005}
}

@inproceedings{cimurs2017bezier,
  title={Bezier curve-based smoothing for path planner with curvature constraint},
  author={Cimurs, Reinis and Hwang, Jaepyung and Suh, Il Hong},
  booktitle={2017 First IEEE International Conference on Robotic Computing (IRC)},
  pages={241--248},
  year={2017},
  organization={IEEE}
}

@inproceedings{bhardwaj2017learning,
  title={Learning heuristic search via imitation},
  author={Bhardwaj, Mohak and Choudhury, Sanjiban and Scherer, Sebastian},
  booktitle={Conference on Robot Learning},
  pages={271--280},
  year={2017},
  organization={PMLR}
}

@article{ahmed2021sit,
  title={Sit: Self-supervised vision transformer},
  author={Ahmed, Sara Atito Ali and Awais, Muhammad and Kittler, Josef},
  journal={CoRR, abs/2104.03602},
  volume={4},
  year={2021}
}

@article{zhu2024vision,
  title={Vision mamba: Efficient visual representation learning with bidirectional state space model},
  author={Zhu, Lianghui and Liao, Bencheng and Zhang, Qian and Wang, Xinlong and Liu, Wenyu and Wang, Xinggang},
  journal={arXiv preprint arXiv:2401.09417},
  year={2024}
}

@article{loshchilov2017decoupled,
  title={Decoupled weight decay regularization},
  author={Loshchilov, Ilya and Hutter, Frank},
  journal={arXiv preprint arXiv:1711.05101},
  year={2017}
}

@article{pohl1970heuristic,
  title={Heuristic search viewed as path finding in a graph},
  author={Pohl, Ira},
  journal={Artificial intelligence},
  volume={1},
  number={3-4},
  pages={193--204},
  year={1970},
  publisher={Elsevier}
}

@article{choudhury2018data,
  title={Data-driven planning via imitation learning},
  author={Choudhury, Sanjiban and Bhardwaj, Mohak and Arora, Sankalp and Kapoor, Ashish and Ranade, Gireeja and Scherer, Sebastian and Dey, Debadeepta},
  journal={The International Journal of Robotics Research},
  volume={37},
  number={13-14},
  pages={1632--1672},
  year={2018},
  publisher={SAGE Publications Sage UK: London, England}
}

@article{sara2019image,
  title={Image quality assessment through FSIM, SSIM, MSE and PSNR—a comparative study},
  author={Sara, Umme and Akter, Morium and Uddin, Mohammad Shorif},
  journal={Journal of Computer and Communications},
  volume={7},
  number={3},
  pages={8--18},
  year={2019},
  publisher={Scientific Research Publishing}
}

@article{simonyan2014very,
  title={Very deep convolutional networks for large-scale image recognition},
  author={Simonyan, Karen and Zisserman, Andrew},
  journal={arXiv preprint arXiv:1409.1556},
  year={2014}
}

@inproceedings{deng2009imagenet,
  title={Imagenet: A large-scale hierarchical image database},
  author={Deng, Jia and Dong, Wei and Socher, Richard and Li, Li-Jia and Li, Kai and Fei-Fei, Li},
  booktitle={2009 IEEE conference on computer vision and pattern recognition},
  pages={248--255},
  year={2009},
  organization={Ieee}
}

@inproceedings{han2023mobile,
  title={Mobile robot path planning based on improved A* algorithm},
  author={Han, Chengyang and Li, Baoying},
  booktitle={2023 IEEE 11th Joint International Information Technology and Artificial Intelligence Conference (ITAIC)},
  volume={11},
  pages={672--676},
  year={2023},
  organization={IEEE}
}

@inproceedings{zhang2024agv,
  title={AGV path planning based on improved A-star algorithm},
  author={Zhang, Dengxing and Chen, Chen and Zhang, Guanyu},
  booktitle={2024 IEEE 7th Advanced Information Technology, Electronic and Automation Control Conference},
  volume={7},
  pages={1590--1595},
  year={2024},
  organization={IEEE}
}

@article{vlastelica2019differentiation,
  title={Differentiation of blackbox combinatorial solvers},
  author={Vlastelica, Marin and Paulus, Anselm and Musil, V{\'\i}t and Martius, Georg and Rol{\'\i}nek, Michal},
  journal={arXiv preprint arXiv:1912.02175},
  year={2019}
}

@article{andreychuk2022multi,
  title={Multi-agent pathfinding with continuous time},
  author={Andreychuk, Anton and Yakovlev, Konstantin and Surynek, Pavel and Atzmon, Dor and Stern, Roni},
  journal={Artificial Intelligence},
  volume={305},
  pages={103662},
  year={2022},
  publisher={Elsevier}
}

@inproceedings{liu2021policy,
  title={Policy learning with constraints in model-free reinforcement learning: A survey},
  author={Liu, Yongshuai and Halev, Avishai and Liu, Xin},
  booktitle={The 30th international joint conference on artificial intelligence (ijcai)},
  year={2021}
}

@article{foead2021systematic,
  title={A systematic literature review of A* pathfinding},
  author={Foead, Daniel and Ghifari, Alifio and Kusuma, Marchel Budi and Hanafiah, Novita and Gunawan, Eric},
  journal={Procedia Computer Science},
  volume={179},
  pages={507--514},
  year={2021},
  publisher={Elsevier}
}

@inproceedings{qiao2023end,
  title={End-to-end vectorized hd-map construction with piecewise bezier curve},
  author={Qiao, Limeng and Ding, Wenjie and Qiu, Xi and Zhang, Chi},
  booktitle={Proceedings of the IEEE/CVF Conference on Computer Vision and Pattern Recognition},
  pages={13218--13228},
  year={2023}
}

@inproceedings{kolmogorov1957representation,
  title={On the representation of continuous functions of many variables by superposition of continuous functions of one variable and addition},
  author={Kolmogorov, Andrei Nikolaevich},
  booktitle={Doklady Akademii Nauk},
  volume={114},
  number={5},
  year={1957},
  organization={Russian Academy of Sciences}
}

@article{yu2024kan,
  title={Kan or mlp: A fairer comparison},
  author={Yu, Runpeng and Yu, Weihao and Wang, Xinchao},
  journal={arXiv preprint arXiv:2407.16674},
  year={2024}
}

@article{li2024u,
  title={U-KAN Makes Strong Backbone for Medical Image Segmentation and Generation},
  author={Li, Chenxin and Liu, Xinyu and Li, Wuyang and Wang, Cheng and Liu, Hengyu and Yuan, Yixuan},
  journal={arXiv preprint arXiv:2406.02918},
  year={2024}
}

@article{vaca2024kolmogorov,
  title={Kolmogorov-arnold networks (kans) for time series analysis},
  author={Vaca-Rubio, Cristian J and Blanco, Luis and Pereira, Roberto and Caus, M{\`a}rius},
  journal={arXiv preprint arXiv:2405.08790},
  year={2024}
}

@article{bao2021beit,
  title={BEiT: BERT Pre-Training of Image Transformers},
  author={Bao, Hangbo and Dong, Li and Wei, Furu},
  journal={arXiv preprint arXiv:2106.08254},
  year={2021}
}

@article{he2021masked,
  title={Masked Autoencoders Are Scalable Vision Learners},
  author={He, Kaiming and Chen, Xinlei and Xie, Saining and Li, Yanghao and Doll{\'a}r, Piotr and Girshick, Ross},
  journal={arXiv preprint arXiv:2111.06377},
  year={2021}
}

@article{lin2024path,
  title={Path Safety Inflation Planning Algorithm Based on Improved JPS},
  author={Lin, Zirong and others},
  journal={The Frontiers of Society, Science and Technology},
  volume={6},
  number={4},
  year={2024},
  publisher={Francis Academic Press}
}

@article{dolgov2010path,
  title={Path planning for autonomous vehicles in unknown semi-structured environments},
  author={Dolgov, Dmitri and Thrun, Sebastian and Montemerlo, Michael and Diebel, James},
  journal={The international journal of robotics research},
  volume={29},
  number={5},
  pages={485--501},
  year={2010},
  publisher={SAGE Publications Sage UK: London, England}
}

@inproceedings{ruml2007best,
  title={Best-First Utility-Guided Search.},
  author={Ruml, Wheeler and Do, Minh Binh},
  booktitle={IJCAI},
  pages={2378--2384},
  year={2007}
}

@inproceedings{veerapaneni2023learning,
  title={Learning local heuristics for search-based navigation planning},
  author={Veerapaneni, Rishi and Saleem, Muhammad Suhail and Likhachev, Maxim},
  booktitle={Proceedings of the International Conference on Automated Planning and Scheduling},
  volume={33},
  number={1},
  pages={634--638},
  year={2023}
}

@article{jiang2020supervised,
  title={Supervised machine learning: a brief primer},
  author={Jiang, Tammy and Gradus, Jaimie L and Rosellini, Anthony J},
  journal={Behavior therapy},
  volume={51},
  number={5},
  pages={675--687},
  year={2020},
  publisher={Elsevier}
}

@incollection{cunningham2008supervised,
  title={Supervised learning},
  author={Cunningham, P{\'a}draig and Cord, Matthieu and Delany, Sarah Jane},
  booktitle={Machine learning techniques for multimedia: case studies on organization and retrieval},
  pages={21--49},
  year={2008},
  publisher={Springer}
}

@article{mehrabian2024implicit,
  title={Implicit Neural Representations with Fourier Kolmogorov-Arnold Networks},
  author={Mehrabian, Ali and Adi, Parsa Mojarad and Heidari, Moein and Hacihaliloglu, Ilker},
  journal={arXiv preprint arXiv:2409.09323},
  year={2024}
}

@inproceedings{kich2024kolmogorov,
  title={Kolmogorov-Arnold Networks for Online Reinforcement Learning},
  author={Kich, Victor A and Bottega, Jair A and Steinmetz, Raul and Grando, Ricardo B and Yorozu, Ayano and Ohya, Akihisa},
  booktitle={2024 24th International Conference on Control, Automation and Systems (ICCAS)},
  pages={958--963},
  year={2024},
  organization={IEEE}
}

@inproceedings{xie2022simmim,
  title={Simmim: A simple framework for masked image modeling},
  author={Xie, Zhenda and Zhang, Zheng and Cao, Yue and Lin, Yutong and Bao, Jianmin and Yao, Zhuliang and Dai, Qi and Hu, Han},
  booktitle={Proceedings of the IEEE/CVF conference on computer vision and pattern recognition},
  pages={9653--9663},
  year={2022}
}

@inproceedings{dong2023peco,
  title={Peco: Perceptual codebook for bert pre-training of vision transformers},
  author={Dong, Xiaoyi and Bao, Jianmin and Zhang, Ting and Chen, Dongdong and Zhang, Weiming and Yuan, Lu and Chen, Dong and Wen, Fang and Yu, Nenghai and Guo, Baining},
  booktitle={Proceedings of the AAAI Conference on Artificial Intelligence},
  volume={37},
  number={1},
  pages={552--560},
  year={2023}
}

@article{numeroso2022learning,
  title={Learning heuristics for A},
  author={Numeroso, Danilo and Bacciu, Davide and Veli{\v{c}}kovi{\'c}, Petar},
  journal={arXiv preprint arXiv:2204.08938},
  year={2022}
}

@article{li2023masked,
  title={Masked modeling for self-supervised representation learning on vision and beyond},
  author={Li, Siyuan and Zhang, Luyuan and Wang, Zedong and Wu, Di and Wu, Lirong and Liu, Zicheng and Xia, Jun and Tan, Cheng and Liu, Yang and Sun, Baigui and others},
  journal={arXiv preprint arXiv:2401.00897},
  year={2023}
}

@article{almuqati2024challenges,
  title={Challenges in Supervised and Unsupervised Learning: A Comprehensive Overview.},
  author={Almuqati, Mohammed Tuays and Sidi, Fatimah and Mohd Rum, Siti Nurulain and Zolkepli, Maslina and Ishak, Iskandar},
  journal={International Journal on Advanced Science, Engineering \& Information Technology},
  volume={14},
  number={4},
  year={2024}
}

@inproceedings{wang2024rs2g,
  title={Rs2g: Data-driven scene-graph extraction and embedding for robust autonomous perception and scenario understanding},
  author={Wang, Junyao and Malawade, Arnav Vaibhav and Zhou, Junhong and Yu, Shih-Yuan and Al Faruque, Mohammad Abdullah},
  booktitle={2024 IEEE/CVF Winter Conference on Applications of Computer Vision (WACV)},
  pages={7478--7487},
  year={2024},
  organization={IEEE}
}

@article{wang2026cruise,
  title={CRUISE: Vision-Language Model-Guided Uncertainty-Aware Cross-Modal Sensor Fusion for Robust Autonomous Driving},
  author={Wang, Junyao and Xu, Yulin and Li, Yu and Khargonekar, Pramod and Faruque, Mohammad Abdullah Al},
  journal={arXiv preprint arXiv:2608.09202},
  year={2026}
}

@article{chen2025ia,
  title={iA*: Imperative learning-based A* search for path planning},
  author={Chen, Xiangyu and Yang, Fan and Wang, Chen},
  journal={IEEE Robotics and Automation Letters},
  year={2025},
  publisher={IEEE}
}

@article{kim2026flexpath,
  title={FlexPath: Adapting Learned Connectivity Guidance to Path Preferences},
  author={Kim, Taehyoung and Schoenbrod, Tim and Eckel, David and Mee{\ss}, Henri},
  journal={arXiv preprint arXiv:2606.10167},
  year={2026}
}

@inproceedings{xu2025daa,
  title={DAA*: Deep Angular A Star for Image-based Path Planning},
  author={Xu, Zhiwei},
  booktitle={Proceedings of the IEEE/CVF International Conference on Computer Vision},
  year={2025}
}

@article{dylan2024convolutional,
  title={Convolutional Kolmogorov-Arnold Networks},
  author={Dylan Bodner, Alexander and Santiago Tepsich, Antonio and Natan Spolski, Jack and Pourteau, Santiago},
  journal={arXiv e-prints},
  pages={arXiv--2406},
  year={2024}
}

@article{hondru2025masked,
  title={Masked image modeling: A survey},
  author={Hondru, Vlad and Croitoru, Florinel Alin and Minaee, Shervin and Ionescu, Radu Tudor and Sebe, Nicu},
  journal={International Journal of Computer Vision},
  volume={133},
  number={10},
  pages={7154--7200},
  year={2025},
  publisher={Springer}
}

@inproceedings{yang2025kolmogorov,
  title={Kolmogorov-arnold transformer},
  author={Yang, Xingyi and Wang, Xinchao},
  booktitle={International Conference on Learning Representations},
  volume={2025},
  pages={76063--76086},
  year={2025}
}

\end{document}